# MULTI-GRANULARITY REPRESENTATION LEARNING FOR SKETCH-BASED DYNAMIC FACE IMAGE RETRIEVAL

*Liang Wang, Dawei Dai\*, Shiyu Fu, Guoyin Wang*

{Key Laboratory of Big Data Intelligent Computing; School of Computer Science and Technology} @Chongqing University of Posts and Telecommunications, Chongqing, China.

## ABSTRACT

In specific scenarios, face sketch can be used to identify a person. However, drawing a face sketch often requires exceptional skill and is time-consuming, limiting its widespread applications in actual scenarios. The new framework of sketch less face image retrieval (SLFIR)[1] attempts to overcome the barriers by providing a means for humans and machines to interact during the drawing process. Considering SLFIR problem, there is a large gap between a partial sketch with few strokes and any whole face photo, resulting in poor performance at the early stages. In this study, we propose a multigranularity (MG) representation learning (**MGRL**) method to address the SLFIR problem, in which we learn the representation of different granularity regions for a partial sketch, and then, by combining all MG regions of the sketches and images, the final distance was determined. In the experiments, our method outperformed state-of-the-art baselines in terms of early retrieval on two accessible datasets. Codes are available at https://github.com/ddw2AIGROUP2CQUPT/MGRL

*Index Terms*—SLFIR; Image Retrieval; Multi-granularity; Partial Sketch;

## 1. INTRODUCTION

Face is the most common biometric used to identify a person. Owing to the requirements of certain practical applications, cross-domain face retrieval based on sketches has become an important task. For example, law enforcement agencies frequently use sketches of visual descriptions provided by onlookers to identify suspects from a database. In addition, with the rapid proliferation of various electronic touchscreen devices, more convenient hand-painted input conditions have been provided for most users, which also have broad application prospects in daily life.

Since, SLFIR belongs to Fine-grained sketch-based image retrieval (FG-SBIR) problem [2] that addresses the problem of retrieving a particular image from a given query sketch. For such problem, manual feature-based approaches exist to perform retrieval by designing an invariant feature descriptor for cross-domain images[3]. Moreover, some methods first convert the cross-domain images into the same style and then perform a retrieval task [7]. Further, some methods aim to learn an efficient function to map sketches and images into a shared embedding space, in which we can directly calculate the similarity between sketches and images [13]. Although significant progress has been made in the FG-SBIR field, such framework still faces challenges in the practice. **This is because it strongly assumes that a high-quality face sketch is ready for retrieval.** However, because most users cannot finish a high-quality face sketch, this prevents the FG-SBIR framework from being widely used in practice.

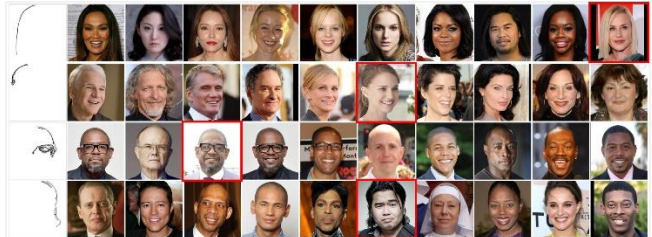

**Fig. 1**: Demonstration of our method on SLFIR problem. Target face photo can be correctly retrieved in Top-10 list with only few strokes.

To break the barriers, our study considers the SLFIR [1] framework, in which retrieval occurs after each stroke, it aims to retrieve the target face image using as few strokes as possible (See **Fig. 1**). This framework can provide a form of human interaction that has considerable potential for sketch-based face image retrieval. Since, a partial sketch with few strokes can show great difference among painters [1]. Therefore, matching the target images of the partial sketch is difficult. In this study, we introduced the idea of multi-granularity (MG) [17] to learn the representations for different granularity regions and then calculated the final distance between the partial sketch and face image by combining multi-granularity regions. Experiments show that our method outperformed state-of-the-art baseline methods in terms of early retrieval efficiency on two publicly accessible datasets.

## 2. METHODOLOGY

### 2.1. Overview

We proposed the **MGRL** method that learn the representation for MG regions of sketch and image to address the SLFIR problem to address the big gap between the partial sketches and the whole images. We aim to retrieve the target facial image using a partial sketch with as few strokes as possible (see **Fig. 1 and 2**, examples of the proposed method). An overview of the proposed method is shown in **Fig. 3**.

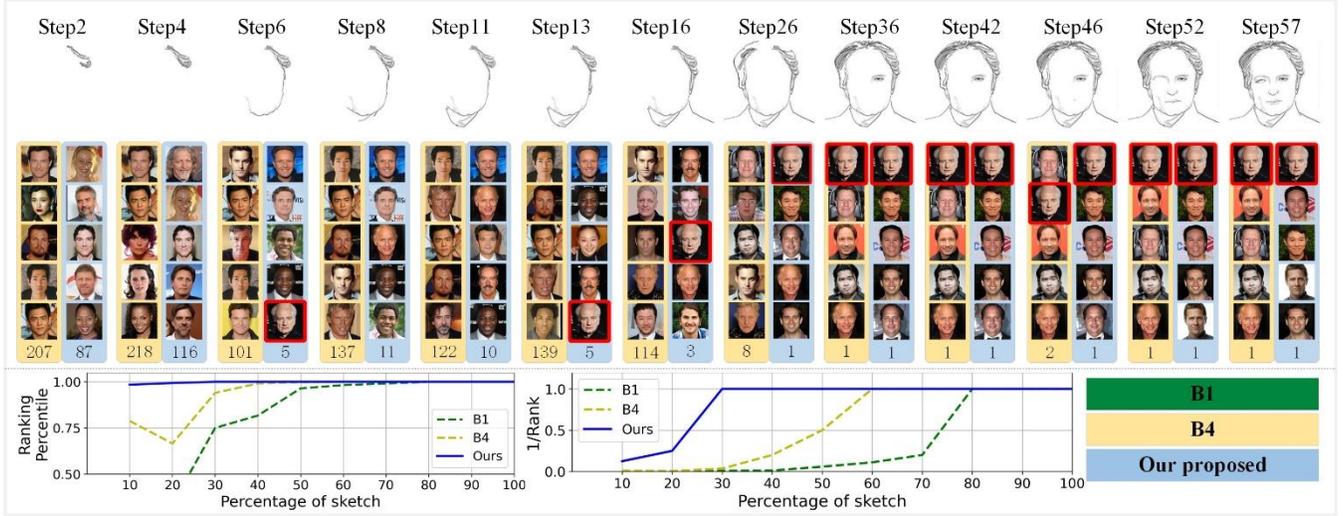

**Fig. 2:** Our proposed MGRL model can retrieve the target face photo using fewer strokes than that of baselines; The number at the bottom denotes the paired (true match) photo's rank at every stage.

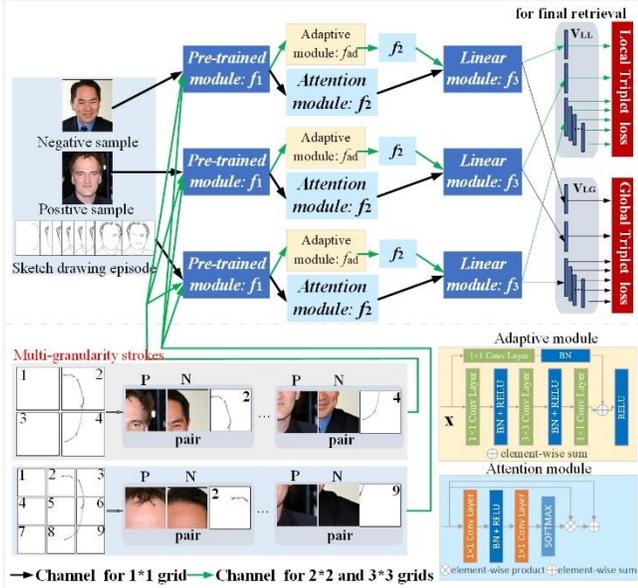

**Fig. 3**: Overview of our approach.

Formally, a sequence of sketches (from the first stroke to the last) is defined as $S = s_1, s_2, ..., s_q$, where $q$ indicates that a complete face sketch contains q strokes. Our model learns an embedding function $F(\cdot) : I \to R^D$ that maps all MG regions of sketch and image to the d-dimensional feature vectors for the final retrieval, that is, we obtain a list of vectors $V_{d-dim} = F(x_{ji})$, $j = 1, ..., m$; $i = 1, ..., n$ for each region from a given gallery of n images, term m indicates the number of regions. For a given query of partial sketch $s$, we obtain the embedding vector of each region using the proposed method, and obtain the **top-k** retrieved images based on the pairwise distance metric. If the target image first appears in the **top-k** list at the current stroke, we consider the **top-k** accuracy true for that sketch.

### 2.2. Backbone Network

We employ a state-of-the-art Triplet network comprising three CNN model branches with shared weights corresponding to positive images, sketches, and negative images. Each branch was divided into three parts. The first part is the pre-trained model $f_1$, which is used to extract features from the input image (sketches, positive and negative images). The Inception-Net model trained on ImageNet was used as the pre-trained model, $Z = f_1(x)$, where $x$ and $Z$ denote the input image and the feature maps. The second part, $f_2$ (Eq. (1)), uses a spatial attention mechanism [18] to learn the feature vectors of global regions of the sketch sequence and its target image, where $f_{att}()$ denotes the attention module. In the third part, $f_3$ (Eq. (2)), we use a simple linear mapping, where A represents the linear mapping used to regulate the dimensionality of the high-dimensional vectors.

$$V_H = f_2(Z) = Gp(Z + Z * f_{att}(Z)) \quad (1)$$

$$V_L = f_3(V_H) = AV_H \quad (2)$$

### 2.3. Adaptive module

Since, the sketch at early drawing process can show great difference among painters, we considered that it was unreasonable to match a partial sketch that contains few strokes with the complete face photos. Especial at the early stage of drawing process, we proposed a Multi-Granularity (MG) representation learning and matching strategy, in which we need to learn the representation for each region in an image. In our model, we first divided an image into $1 \times 1$, $2 \times 2$, and $3 \times 3$

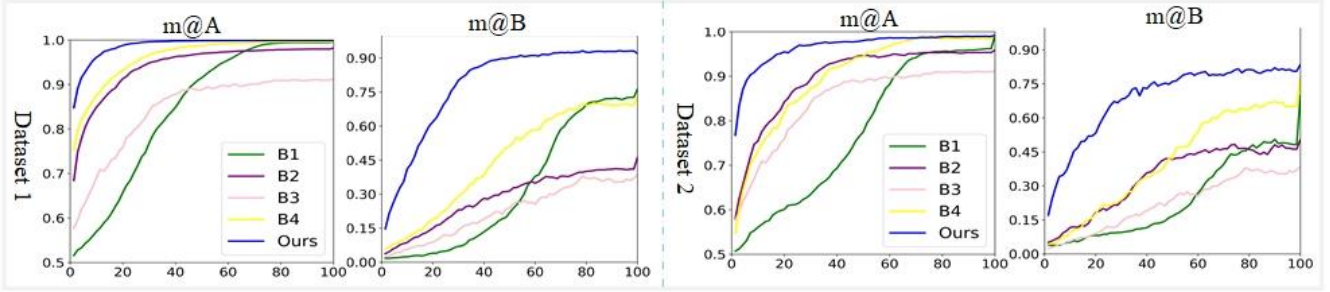

Fig. 4: Performance of early retrieval. We visualized it using each percentage of the sketch. A higher value indicates better early retrieval performance.

$$D(x_{ph}, x_s) = \alpha * Ed(V_{SL1*1}, V_{ph1*1}) + \beta * (\frac{1}{k_2} * \sum_{i=0}^{k_2} Ed(V_{SL2*2}[i], V_{ph2*2}[i])) + \gamma * (\frac{1}{k_3} * \sum_{i=0}^{k_3} Ed(V_{SL3*3}[i], V_{ph3*3}[i])) \quad (3)$$

grid regions to obtain regions with different granularities, each regarded as one sample. Subsequently, we adopt a residual block as an adaptive module attached to the backbone network to address the fine-granularity regions (as shown in **Fig. 3**). The requirement for the additional module is summarized as follows: the finer the region, the smaller the size; if there is no adaptive module, it would be necessary to enlarge each subregion, which significantly increases the computational complexity.

### 2.4. Multi-granularity Representation Learning

As shown in **Fig. 3**, we employed a simple linear layer following the attention module to map high-dimensional vectors to low-dimensional vectors for the final retrieval. During the training process, we first fixed f1 and updated the other parts. Each partial sketch in the sketch-drawing episode was divided into $1 \times 1$, $2 \times 2$, and $3 \times 3$ grid regions, and the grid regions with little or no information was eliminated. We designed a triplet loss to train the model to minimize (maximize) the distance between the sketch and the positive (negative) sample, considering the corresponding MG regions of the face sketch and image. As shown in Eq. (4), the terms $v_{SL}$, $v_p$ and $v_n$ indicate the final representations of the sketch and the positive and negative samples, respectively.

$$J_1 = max(\sum_{i=0}^{q} d(v_{SL}, v_p) - d(v_{SL}, v_n) + \theta, 0) \quad (4)$$

The similarity between one image and one partial sketch was determined by calculating the MG distances. As shown in Eq. (3), $Ed(V_{SL1*1}, V_{ph1*1})$, $Ed(V_{SL2*2}, V_{ph2*2})$, $Ed(V_{SL3*3}, V_{ph3*3})$ denotes the Euclidean distances between the different granularities of the partial sketch and image, where $V_{SL1*1}$($V_{SL2*2}$, $V_{SL3*3}$) and $V_{ph1*1}$($V_{ph2*2}$, $V_{ph3*3}$) represent the low-dimensional vectors of the partial sketch and image obtained from Eq.(2); Term $i$ denotes the $i^{th}$ region of this partial sketch and its corresponding region of an image; $k_1$, $k_2$ and $k_3$ ($k_1$=1, $k_2$=4 and $k_3$=9) denote the number of regions of this sketch that contain a certain amount of stroke information; $\alpha$ and $\beta$, and $\gamma$ represent the weights of three distances. Here, we set $\alpha=\beta=\gamma=1$.

## 3. EXPERIMENTS

### 3.1. Dataset

The FS2K-SDE includes two datasets of face sketch-drawing episodes (**Dataset 1** and **Dataset 2**), which are generated by a cognitive strategy of large-scale priority based on the FS2K, which is a face-sketch synthesis dataset [1]. **Dataset 1** contained 107,030 sketches and 1529 images, of which 75,530 sketches and 1079 images were used for training and the rest for testing. **Dataset 2** contained 33,390 sketches and 477 images, of which 23,380 sketches and 334 images were used for training, and the rest for testing.

### 3.2. Implementation details

Experiments were conducted on a 40GB Nvidia A100 GPU. A triplet loss function with a margin of 0.3 ($\theta$ in Eq. (4)) and the Adam optimizer were used to train the model; For the Inception-V3 part, we fixed the parameters of the first few layers of convolution in the pre-trained network and updated the parameters of the later layer with a learning rate of 5e-4 as the training progressed towards the target task, and updated the residual-like block, attention module, and fully connected layers initialized using the Kaiming normal [19] at a learning rate of 5e-3. We input sketch sequence images and images with a batch size of 32. For $2 \times 2$, $3 \times 3$ granularity regions of image. The proposed model was trained for over 20 epochs.

### 3.3 Evaluation Metric

Regarding the sketch-based image retrieval framework, we prioritized the target face images appeared at the top of the list. Specifically, $m@A$ (ranking percentile) and $m@B$ (1/rank versus percentage of sketches)[15] were used to evaluate the early retrieval performance for partial sketches (average performance of all stages). To better reflect the early retrieval, we also employed *w@mA* and *w@mB* [1]. *A@5* is traditional top-5 accuracy based on complete sketch.

Table 1: Comparation with 16-dimension feature-embedding.

| | Dataset 1 | | | | | Dataset 2 | | | | |
|---|---|---|---|---|---|---|---|---|---|---|
| | m@A | m@B | w@mA | w@mB | A@5 | m@A | m@B | w@mA | w@mB | A@5 |
| B1[20](2017) | 84.77 | 32.69 | 50.40 | 15.83 | 91.33 | 77.83 | 24.59 | 46.00 | 12.47 | 94.41 |
| B2[20](2017) | 94.16 | 28.58 | 58.18 | 15.83 | 64.22 | 89.77 | 34.14 | 54.99 | 18.96 | 69.23 |
| B3[21](2020) | 84.42 | 22.76 | 51.52 | 12.21 | 51.78 | 85.65 | 26.69 | 51.91 | 14.59 | 60.23 |
| B4[1](2023) | 96.22 | 45.48 | 59.57 | 24.56 | 90.00 | 90.22 | 41.55 | 54.85 | 22.22 | **95.80** |
| Ours | **98.80** | **78.92** | **61.69** | **46.20** | **97.11** | **96.65** | **69.19** | **60.12** | **40.80** | 95.10 |

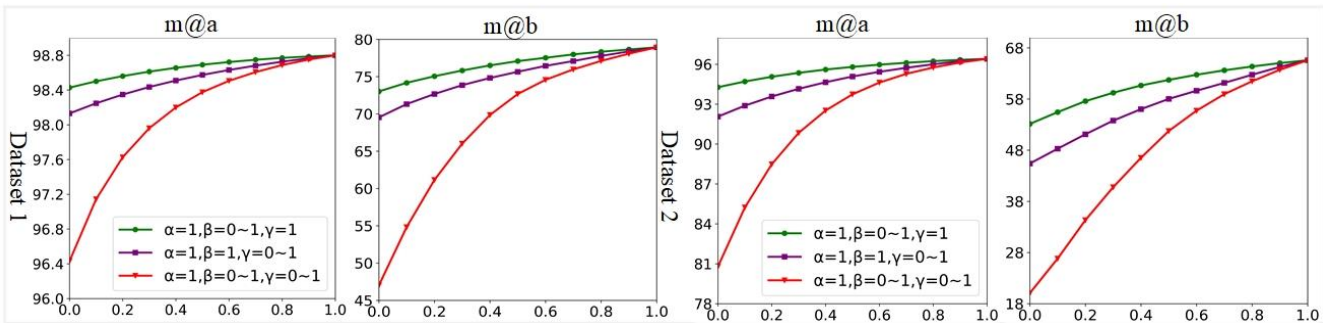

**Fig. 5**: Ablation experiments. Retrieval performance when fixing one or two weight and changing another from 0-1. A higher value indicates better retrieval performance.

### 3.4. Baselines

**B1** [20]: A Classical FG-SBIR model was trained with triplet loss using only the complete face sketches. The framework of **B2**[20] was the same as that of B1, except that we trained B2 using the sketch-drawing episode data. **B3**(RL-based) [21]: A RL-based method was developed to optimize the representations of the partial sketches obtained from the CNN model. **B4** [1]: A CNN+LSTM model was developed to optimize the sequence of partial sketches.

### 3.5. Performance Analysis

The performance of early retrieval of our proposed is compared with that of the baselines, as shown in **Fig. 4**. We can note the following: (1) Such new framework can achieve retrieving the target face photo using a partial face sketch with fewer strokes; (2) Our proposed method performs much better at early retrieval than that of baselines without decreasing top-5 accuracy. All the quantitative results are shown in Table.1, and it can be concluded that our proposed method is significantly superior to all baselines in early retrieval performance. As shown in Table.2, our proposed model performs considerably better than B4 for each dimension.

#### 3.5.1. Ablation analysis

In this section, we analyze the influence of MG representations on the early retrieval performance, as shown in **Fig. 5**, which can be operated by adjusting the weights ($\beta$ and $\gamma$) of MG distances. In Eq. (3), $\beta$ and $\gamma$ indicate the weights of the $2 \times 2$ and $3 \times 3$ granularity respectively. Here, we fixed $\alpha =1$, $\beta = 1$ ($\gamma = 1$) and adjusted the value of $\gamma$ ($\beta$) from 0 to 1 or we fixed $\alpha =1$ and adjusted the value of $\beta$ and $\gamma$ from 0 to 1. As shown in **Fig. 5**, we observe that when fixing one or two weight and gradually increasing another, the early retrieval performance gradually improves. Experiments verified that our MG representation can significantly improve the early performance of SLFIR problem.

**Table 2:** Comparation with varying feature-embedding.

| | | B4[1](2023) | | Ours | |
|---|---|---|---|---|---|
| | Dim | w@mA | w@mB | w@mA | w@mB |
| Dataset 1 | 8 | 59.34 | 23.00 | **61.35** | **41.31** |
| | 16 | 59.57 | 24.56 | **61.69** | **46.20** |
| | 32 | 59.33 | 22.13 | **61.62** | **46.23** |
| | 64 | 58.63 | 19.81 | **61.75** | **45.17** |
| Dataset 2 | 8 | 55.52 | 23.17 | **59.96** | **37.68** |
| | 16 | 54.85 | 22.22 | **60.12** | **40.80** |
| | 32 | 53.43 | 21.36 | **60.83** | **45.27** |
| | 64 | 53.13 | 18.73 | **60.72** | **43.85** |

### 4. CONCLUSION AND FUTURE WORK

Drawing a face sketch is time-consuming and requires exceptional skills, which make sketch-based face image retrieval inefficient. SLFIR framework aims to break these limitations. Considering the SLFIR problem, we propose a MGRL method to address the diversity of early sketching, and experiments verify that our method can improve the early performance significantly. However, this novel framework still encounters challenges, for example, (1) how to select a stroke that will be conducive to retrieval, and (2) how to make better use of the interaction between human and machine.